\def\year{2021}\relax
\documentclass[letterpaper]{article} % DO NOT CHANGE THIS
\usepackage{aaai21}  % DO NOT CHANGE THIS
\usepackage{times}  % DO NOT CHANGE THIS
\usepackage{helvet} % DO NOT CHANGE THIS
\usepackage{courier}  % DO NOT CHANGE THIS
\usepackage[hyphens]{url}  % DO NOT CHANGE THIS
\usepackage{graphicx} % DO NOT CHANGE THIS
\usepackage{natbib}  % DO NOT CHANGE THIS AND DO NOT ADD ANY OPTIONS TO IT
\usepackage{caption} % DO NOT CHANGE THIS AND DO NOT ADD ANY OPTIONS TO IT
\usepackage{colortbl}
\usepackage{amsmath}
\usepackage{multirow}
\usepackage{hyperref}
\usepackage{booktabs}

\newcommand{\conceptnet}{\texttt{ConceptNet}}
\newcommand{\wordnet}{\texttt{WordNet}}
\newcommand{\wikidata}{\texttt{Wikidata}}
\newcommand{\visualgenome}{\texttt{VisualGenome}}
\newcommand{\commongen}{\texttt{CommonGen}}
\newcommand{\atomic}{\texttt{ATOMIC}}
\newcommand{\dice}{\texttt{DICE}}
\newcommand{\senticnet}{\texttt{SenticNet}}

\newcommand{\paperidvar}{10}

\begin{document}
%\title{Commonsense Knowledge Extraction and Injection methods for lexically constrained text generation }
\title{Lexically-constrained Text Generation through \\ Commonsense Knowledge Extraction and Injection}

% ANONYMOUS submission author block
\iffalse % comment out for ANONYMOUS submission
\author{Anonymous AAAI 2021 Workshop Submission \\}
\affiliations{Paper ID \paperidvar}
\fi % comment out for ANONYMOUS submission

% CAMERA READY author block
%\iffalse % comment out for CAMERA READY version
\author{
Yikang Li\textsuperscript{\rm 1}\thanks{Correspondence},
Pulkit Goel\textsuperscript{\rm 1}$^*$,
Varsha Kuppur Rajendra\textsuperscript{\rm 1},
Har Simrat Singh\textsuperscript{\rm 1},\\
Jonathan Francis\textsuperscript{\rm 1,2}$^*$,
Kaixin Ma\textsuperscript{\rm 1}$^*$,
Eric Nyberg\textsuperscript{\rm 1},
Alessandro Oltramari\textsuperscript{\rm 2}\\
}
\affiliations{
\textsuperscript{\rm 1}Language Technologies Institute, School of Computer Science, Carnegie Mellon University\\
\textsuperscript{\rm 2}Human-Machine Collaboration, Bosch Research Pittsburgh\\
\{yikangli, pulkitgo, vkuppurr, harsimrs, jmf1, kaixinm, ehn\}@cs.cmu.edu, alessando.oltramari@us.bosch.com
\vspace{2mm}
}
%\fi % comment out for CAMERA READY version

%\author{Bosch\\Association for the Advancement of Artificial Intelligence\\ 2275 East Bayshore Road, Suite 160\\ Palo Alto, California 94303\\
%}
\maketitle

%%%%%%%%%%%%%%%%%%%%%%%%%%%%%%%%%%%%%%%%%%%%%%%
%%%%%%%%%%%%%%%%%%%%%%%%%%%%%%%%%%%%%%%%%%%%%%%

\begin{abstract}
\begin{quote}
%Generative Modelling for Natural Language Generation has been a classical problem in Natural Language Processing that is yet to achieve human level performance. 
Conditional text generation has been a challenging task that is yet to see human-level performance from state-of-the-art models. In this work, we specifically focus on the \commongen~benchmark, wherein the aim is to generate a plausible sentence for a given set of input concepts. Despite advances in other tasks, large pre-trained language models that are fine-tuned on this dataset often produce sentences that are syntactically correct but qualitatively deviate from a human understanding of common sense. Furthermore, generated sequences are unable to fulfill such lexical requirements as matching part-of-speech and full concept coverage. In this paper, we explore how commonsense knowledge graphs can enhance model performance, with respect to commonsense reasoning and lexically-constrained decoding. We propose strategies for enhancing the semantic correctness of the generated text, which we accomplish through: extracting commonsense relations from \conceptnet, injecting these relations into the Unified Language Model (UniLM) through attention mechanisms, and enforcing the aforementioned lexical requirements through output constraints. By performing several ablations, we find that commonsense injection enables the generation of sentences that are more aligned with human understanding, while remaining compliant with lexical requirements.
\end{quote}
\end{abstract}

%  of lexically-constrained text generation
% Further, these generations are often unable to fulfill all the hard lexical constraints. 
% lexical constraints - by modifying beam search during decoding; mention this in intro, instead

% \noindent\jf{Re: Abstract: Specify the challenges in the field and with other models; allude to the improvements that we make over SOTA and how; allude to our performance}\\
%\noindent\jf{Captions go above tables and below figures}\\
%\noindent\jf{\textit{General comment:} We should unify the format of all of our tables. I recommend using the standard ICML / ICLR / NeurIPS format (\texttt{toprule}, \texttt{bottomrule}, \texttt{hline}s or \texttt{midrule}s in between, no column separation bars), from the \texttt{booktabs} package.}\\

\noindent 

%%%%%%%%%%%%%%%%%%%%%%%%%%%%%%%%%%%%%%%%%%%%%%%
%%%%%%%%%%%%%%%%%%%%%%%%%%%%%%%%%%%%%%%%%%%%%%%

\section{Introduction}
\label{section:introduction}

Natural language generation is the backbone for a plethora of applications, where language models|such as GPT-2 \cite{radford2019language}, UniLM \cite{NEURIPS2019_c20bb2d9}, and BART \cite{lewis2019bart}|are leveraged for challenging tasks, such as dialogue generation, story-telling, text summarization, and descriptive question answering. While these language models have brought about significant performance improvements, largely due to their scale, these models are yet to reach human-level performance. Model performance worsens further for tasks, such as constrained text generation, where the generated text is expected to follow a set of pre-defined rules or ``requirements." Examples of lexically-constrained text generation include natural language generation, conditioned on the content in structured tables, or caption generation, based on a list of words. 

The \commongen~dataset \cite{lin2020commongen} is an instance of the word-conditioned caption generation task. Here, the aim is to generate a syntactically- and semantically-coherent sentence from a given concept-set; these concepts are usually nouns and verbs and represent entities from day-to-day life. The authors observed that, although popular language models generate sentences with reasonable grammatical structure, they struggle with two major aspects of the task. Firstly, generated sentences did not completely adhere to a human's understanding of common sense. As an example, for the a given set of concepts, say \textit{\{dog, catch, throw, frisbee\}}, GPT-2 generated the following sentence: \textit{A dog throws a frisbee at a football player}, while UniLM generated ``\textit{Two dogs are throwing frisbee at each other}". Even though the generated sentences have syntactic integrity, the language models were far from grasping the essence of common sense. Secondly, these pre-trained language models also struggled to include all the given concepts in the concept-set, instead producing sentences with only partial coverage. In the examples, above, the concept \textit{catch} was missing in text generations from both GPT-2 and UniLM. Another example is of text generation from the T5 language model \cite{raffel2019exploring}, where the generated sentence, \textit{dog catches a frisbee and throws it to a dog}, not only lacks common sense but also sees a repetition of the concept \textit{dog}. These phenomena of missing concepts, repetition, and lacking concept coverage limit the quality of generated text and, by extension, negatively affect models' task performance.

We hypothesize that addressing these common faults, directly, will lead to improvements in generated sentence quality and to increases in model performance. In this paper, we address these limitations in two parts. First, we ground the given concept-set on multi-hop knowledge primitives found in existing commonsense knowledge graphs: we bias models by extracting concept-specific relations from knowledge graphs and by injecting this additional context into early layers of popular language model classes. In this way, the model is able to expand the original concept set, to include additional concepts that, ultimately, make the generated text more realistic; through ablation studies, we show that this coupled extraction/injection process increases the semantic correctness of the generated text output. Next, we further enhance the models' adherence to task requirements by imposing lexical constraints on the models' output, through adjusted beam search decoding. In this manner, we significantly reduce issues related to repetitive concept phenomena. Our contributions include: (1) strategies to extract commonsense information from knowledge graphs; (2) attention mechanism to inject the commonsense knowledge into the language model; and (3) enforcing lexical constraints by modifying beam search decoding. Our strategy can thus be formulated as a three step process:
\begin{itemize}
\item \textbf{Knowledge Extraction}: For a given set of concepts, knowledge graphs can help identify relationships between a pair of concepts, as seen in a real-world scenario. In this paper, we focus on strategies to extract relations between concepts from the \conceptnet~knowledge graph.
\item \textbf{Knowledge Injection}: We explore different methodologies that aim to inject the extracted relations between concepts into the language model. We limit our experiments to UniLM, which is a transformer-based seq2seq language model, and had state of the art performance in many metrics on the \commongen~dataset. We discuss both attention-based and non attention-based approaches. 
\item \textbf{Constrained Decoding}: We modify the beam search decoding to 
assess the best-$N$ beams for their adherence to the given lexical-constraints. Sentence with the highest beam score that includes all the given concepts as well as matches the given part-of-speech (POS) tags of the concepts is selected as the output.

\end{itemize}

%%%%%%%%%%%%%%%%%%%%%%%%%%%%%%%%%%%%%%%%%%%%%%%
%%%%%%%%%%%%%%%%%%%%%%%%%%%%%%%%%%%%%%%%%%%%%%%

\section{Related Work}
\label{section:relatedwork}

%\jf{Remember: the related works section is not just a survey of recent works — it needs to be an organised discourse on why contemporary methods (including last year’s HyKAS paper!) are insufficient or why their focus is different from ours in this paper. This way, you can see how the Related Works section serves as a concrete secondary-motivation section, while also foreshadowing our approach and experimental design.}

\subsection{External Knowledge Resources}

Commonsense knowledge graphs seek to codify a human-like understanding of the relations between concepts and events that occur in the real world. Chief among these is \conceptnet~\cite{speer2016conceptnet}, which encodes commonsense knowledge as triples, of the form: $[C_1, r, C_2]$, where $C_1$ and $C_2$ represent commonly-used head and tail concepts and \textit{$r$} denotes the type of relation between these concepts, such as RelatedTo, Synonym, etc. The ATOMIC knowledge graph \cite{sap2019atomic} enables reasoning about \textit{what}, \textit{how}, and \textit{why} a cause can lead to an effect: this resource models the interactions between concepts/entities as \textit{if-then} relationships, as opposed to the taxonomic relations modeled by \conceptnet. The \dice~knowledge graph \cite{chalier2020joint} gives a multi-faceted nature to concept relations by providing scores for characterizing them as plausible, remarkable, salient, and typical. In other words, \dice~comments on the circumstances under which two concepts are related and how. Other knowledge graphs incorporate task-specific commonsense knowledge, such as \senticnet~\cite{Cambria2020SenticNet6E}, which is custom built for concept-level sentiment analysis. Previous work has also consolidated and utilized a diverse set of commonsense knowledge graphs (including \wordnet~\cite{10.1145/219717.219748}, \conceptnet, \atomic, \wikidata~\cite{ilievski2020commonsense}, and \visualgenome~\cite{krishnavisualgenome}) into a unified framework \cite{ilievski2020consolidating, ma2020knowledgedriven}, used for multiple-choice commonsense question answering. Regardless of the task \textit{format}, we follow \citet{ma-etal-2019-towards, ma2020knowledgedriven} in asserting that the choice of external resource plays a significant role in the downstream performance|based on the alignment between the task semantics and the type of common sense encoded by the resource; consequently, we utilize \conceptnet~in this work.

\subsection{Knowledge Manipulation}
%knowledge extraction and injection

Given knowledge primitives from an external resource, various approaches have been proposed for transforming the symbolic knowledge into a neural representation that can be easily consumed by language models. We call this process of infusing models with knowledge as ``knowledge injection." 
Among the earliest works involving extraction and injection of knowledge from a KG are, \cite{ahn2017neural} and \cite{yang-etal-2017-reference} where the authors propose architectures which combine symbolic knowledge provided by a KG with an RNN language model. \citet{lin2020commongen} propose concatenating human-generated hints (``rationale" tokens) to the input concepts, for conditional text generation. 
\citet{bauer2019commonsense, mihaylov-frank-2018-knowledgeable} inject knowledge into the intermediate layers of neural models, but they focus on reading comprehension and multiple-choice question answering tasks, where the role of common sense is less defined. \citet{ma-etal-2019-towards, oltramari2020neurosymbolic, ma2020knowledgedriven, wang2020connecting} propose using attention mechanisms for commonsense knowledge injection, for multiple-choice commonsense question answering, by applying attention with respect to the question followed by an Option Comparison Network (OCN) cell. Inspired by \citet{ma-etal-2019-towards, lin2020commongen}, we adapt and unify these methodologies for lexically-constrained, conditional text generation on commongen. 
Similar to the recent works \cite{lauscher2020common, liu2020commonsense}, we inject \conceptnet~relations in sentence form into the transformer layers. 
While these works use an adapter-based residual bottleneck and evidence generators for NLU tasks like classification or MCQA, respectively, we introduce a multi-linear attention distribution to the hidden representation of the encoder, to solve the problem of constrained text generation. 
Works like \cite{liu2019kbert} and \cite{chen2020kgpt} also adopt a similar strategy for knowledge injection into a language model albeit the problem that they intend to solve is different. 

The aforementioned works focus on the downstream prediction tasks, with less emphasis on analysing the ideal type and size of knowledge to be injected. Our work brings together the techniques of better knowledge extraction, commonsense knowledge manipulation and constrained text generation. %We also experiment with respect to various knowledge extraction methodologies, injection methodologies and decoding methods.

\subsection{Constrained Text Generation}

As a methodology, lexically-constrained text generation enjoys application to many real-world domains, such as dialogue systems, machine translation, and paragraph/story generation. Table-to-text is one such task, where, given a structured dataset such as a table, the aim is to generate human-standard sentence(s) with the constraint that all the words specified in the table should be included in the generated text. In their work on Neural Template generation, \citet{wiseman2018learning} explore conditional text generation on the WikiBio and E2E datasets by learning latent, discrete templates using a neural Hidden Semi-Markov Model (HSMM) decoder; \citet{lebret-etal-2016-neural} propose copy actions for the same task, in order to include all the given concepts in the generated text. In this paper, we focus on sentence generation from a given list of concepts, prescribed by the \commongen~dataset. In their work on this dataset, \citet{lin2020commongen} experiment with multiple baseline models and pre-trained language models, including GPT-2, BERT, UniLM, BART, and T5.

\section{Approach}
\label{section:approach}

\subsection{Knowledge Extraction Methodology}
\label{subsection:extraction}

We refer to ``knowledge extraction" as the process of obtaining relevant knowledge primitives (e.g., triples or multi-hop paths) from an external resource (\conceptnet), in order to satisfy a downstream task (question answering, text generation). In this section, we discuss multi-hop knowledge extraction with \conceptnet, knowledge selection strategies, and query expansion|all geared towards obtaining the best knowledge for downstream use with the model.

\subsubsection{Multi-hop extraction}

We derive our multi-hop commonsense knowledge extraction procedure, as an extension of the single-hop case used in \citet{ma-etal-2019-towards, ma2020knowledgedriven, oltramari2020neurosymbolic}. The \commongen~dataset contains sample concept-sets, with set lengths that range from 3-5 concepts. As a consequence of how the dataset was generated, most concept-sets are connected by either 1-hop or 2-hop relations in \conceptnet~\cite{lin2020commongen}. For a given concept-set, our goal is to extract the commonsense relations that connect the concepts in the concept-set. The multi-hop extraction method is developed as follows: 

\begin{itemize}
\item In order to capture salient relations between concept-set elements, we consider two hops by searching among all 1-hop neighbors of each concept-set element and setting these neighbors as the root concept to look for \textit{their} relations with other concept-set elements. For example, given a concept-set [``broccoli", ``cheese", ``chicken", ``pizza"], we extract 1-hop relations such as [``cheese", ``AtLocation", ``pizza"] and 2-hop relations such as [``broccoli", ``AtLocation", ``plate", ``RelatedTo", ``pizza"]. 

\item Sometimes, there are low-connectivity concepts that have no 1-hop or 2-hop relations with other input concepts. For these, we use 3-hop extractions. However, searching among all possible 3-hop relations is time-consuming. Instead of using all neighbors, we use only the five ``nearest" neighbors (i.e., those with highest \conceptnet~relation weights, as a proxy for the strength by which the edge expresses the assertion) as the root concepts for the second-level search. Because 3-hop relations involve more than 3 components, we use the term ``knowledge relations", in the following sections, to refer to all extracted relations.
\end{itemize}

\subsubsection{Knowledge Selection}
One issue with multi-hop extraction is that the same pair of concepts can be linked by different relations, yielding a noisy inductive bias for training models. In fact, for the \commongen~task, an average of 9 knowledge relations are extracted for each concept-set. However, we recognize that some knowledge relations are more useful than others. While it is hard to automatically evaluate relation relevance, we propose heuristic selection mechanisms.

\begin{itemize}
\item \textit{Relation types and POS-based selection.} Our knowledge extraction process excludes such relation types as: `FormOf', `DerivedFrom', `Antonym' and `DistinctFrom’. `FormOf' and `DerivedFrom' can be discarded, since they indicate purely syntactic relations, as in [`walk', `FormOf', `walking']. `Antonym' and `DistinctFrom' are designed to link concepts that have opposing semantic interpretations, which can harm the goal of linking concept-set elements together. We only extract relations following the given POS tags of concept-set elements. 

%\item  \textit{Length based}: One assumption is that a shorter relation is better than a longer relation. The intuition is that single-hop relations may indicate more tight connections between two concepts. Our knowledge selection mechanism will prefer single-hops over multi-hops. 

%\item  \textit{Frequency based}: More frequently-used relation types appear better than rarely-used relation type. Knowledge paths with more common relation types, such as ``RelatedTo" and ``LocatedAt", are preferred over those with rare types like ``SimilarTo" or ``MannerOf". %This is still just an assumption that we could experiment with.

\item  \textit{Random subset selection}: Instead of filtering on the above-mentioned criteria, we can perform random subset selection over all knowledge relations by assigning a random selection probability to each of the relations and selecting the relations above a particular threshold. The selection can be constrained such that at least one relation for each input concept will be kept.

\item  \textit{Subset selection using Prior probability}: 
\label{section:prior} Instead of assigning a random probability to all knowledge relations, we compute a prior probability over the relation types i.e. the number of occurrences of a relation type over total number of relations. We also have a random component similar to 'random subset selection' so that we retain some of the relations with less frequent relation type. The probability of choosing a knowledge relation is the sum of prior and random probabilities. We observe the distribution obtained on the training data and choose a threshold for relation selection. The average number of relations for each concept-set is now 3 to 4.

\end{itemize}
%Permutations can also be performed since the order in which we send the relations is also important. We can have the model output multiple sentences with different randomizations/permutations and then rank the outputs to select the best one as final output.

\subsubsection{Query Expansion}
Apart from knowledge extraction, we perform query expansion on the given concept-set. The motivation is to make our model generate sentences with more contents by expanding our concept-set. We count the frequency for all single-hop neighbors of input concepts and select neighbors that are connected with more than half of the input concepts. For example, the given concept set [``drill", ``field", ``run", ``team"] has [``baseball", ``sport", ``football"] as expansion concepts. The expanded concepts are obtained in order of their frequencies, so that the number of expanded concepts can be adjusted by setting a threshold. They serve as supplementary concepts to feed into the model.

\subsection{Knowledge Injection Methodology}
\label{subsection:injection}

We use UniLM \cite{dong2019unified} as the backbone architecture, adding improvisations with respect to commonsense injection within it. On a high-level the architecture of this Language Model (LM) is as described in Figure \ref{fig:uniLm-hykas-integration} consisting of multiple stacked layers of bidirectional Transformer encoder and a unidirectional decoder being optimized for Masked LM loss. 
Originally, UniLM takes input in the form \emph{(Concept-Set [SEP] TARGET\_SENT)}. We finetune the UniLM model for the generation task in seq-2-seq mode. Using UniLM as the base network, we attempt to perform language generation requiring commonsense knowledge using the following methods:

\subsubsection{Knowledge Concatenation}
As a baseline injection methodology, we concatenate the extracted tokens from the concept relations to our inputs. This method is based on rationale concatenated methods proposed by \cite{lin2020commongen} in their work. The expanded list of input tokens is then fed to the language model for text generation. 

\subsubsection{Attention Injection Method}

To better handle the amount and context of commonsense knowledge fed to the LM, we adopt attention based injection methodology following HyKAS~\cite{ma-etal-2019-towards}. We extract commonsense knowledge from \conceptnet~as described in the \nameref{subsection:extraction} section and provide that to the model in the form of knowledge relations. 
%The ideal number, type and depth of these relations have been explored in various experiments described later in Section: \nameref{section:experiments}.

\begin{figure}[ht]
\includegraphics[width=0.40\textwidth]{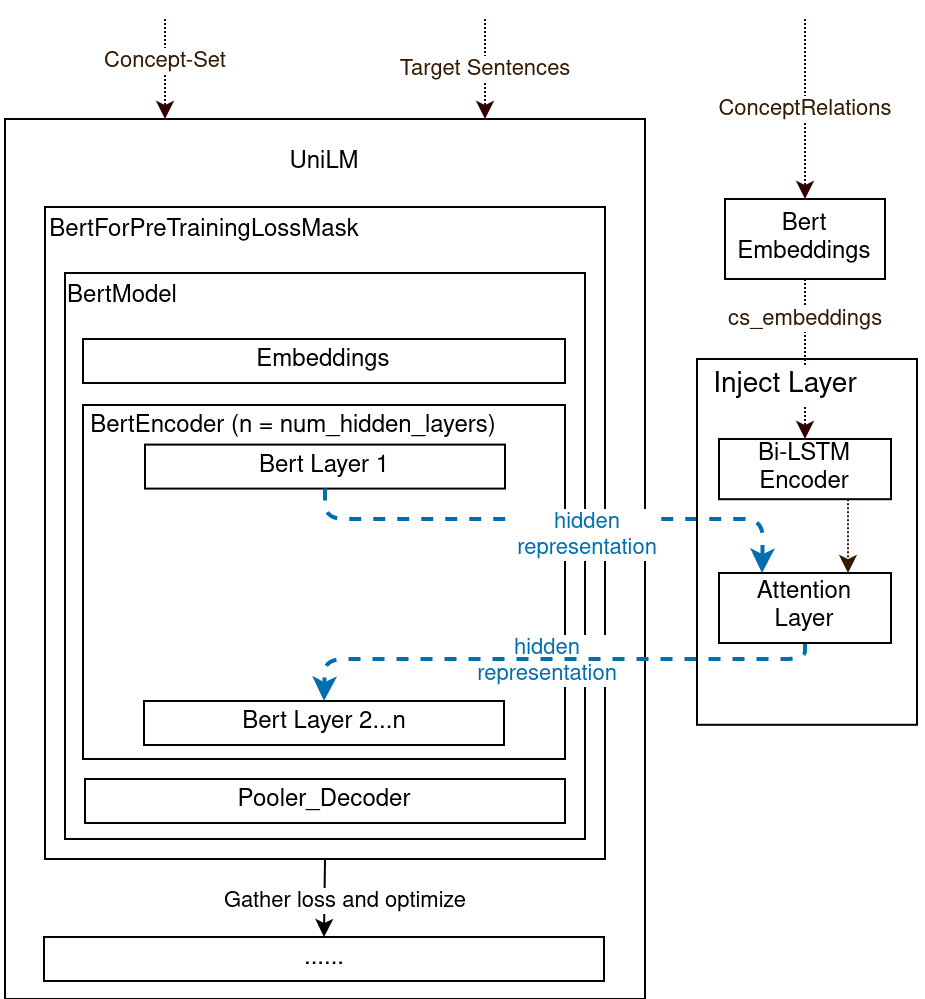}
\centering
\caption{Integrating Attention Injection with UniLM}
\label{fig:uniLm-hykas-integration}
\end{figure}

We make use of commonsense embeddings (marked as \emph{cs\_embeddings} in Figure \ref{fig:uniLm-hykas-integration}) which are sent to a bi-LSTM encoder to get commonsense encodings (\emph{cs$\_$encodings}). We consider the hidden representation of the concept-set segment after the first encoder layer of UniLM and compute Key-Value attention with the \emph{cs$\_$encodings} obtained from the bi-LSTM layer.   We compute attention according to the given equations. Here, Q,K,V are the projections of \emph{cs$\_$encodings}, $H_{hid}$ is the hidden representation from UniLM, M is the joint mask for \emph{commonsense$\_$encoding} and $H_{hid}$: 
\begin{align*}
&A = softmax(QK^T + M)\\
&H_{ctxt} = A \cdot V\\
&H_{attn} = W^T \cdot H_{ctxt} + H_{hid}
\end{align*}

We use the obtained attention\_scores along with the hidden representation to get \emph{commonsense attended hidden representation}, $H_{attn}$. This is inserted back to the UniLM model and it propagates through the consecutive encoder layers.

\begin{itemize}
    \item \textbf{Attention on Knowledge relations} The knowledge relations are given as additional inputs to the LM along with available input set, \emph{[Concept-Set, Target Sentence]}. 
We convert the relations into regular sentences, e.g. \emph{( football $\rightarrow$ RelatedTo $\rightarrow$ sport )} is converted to the form "football related to sport". We further tokenize and obtain BertEmbeddings (cs\_embeddings) for such sentences, which are sent to the encoder described above. We compute attention between these relations and the input concept-net and inject it into the hidden representation of the encoder.
    \item \textbf{Attention for expansion concepts}    
    In order to understand the context required for sentence generation we  derive additional concepts from \conceptnet~which tries to add details to the story (sentence)  being built from concepts in the given concept-set as building blocks. We call this set as the expanded concept-set. The expanded concept-sets are given as input to the LM by concatenating them with provided concept-sets. 
    In a general setting with injection, we try to capture the relation between the given concepts and the concept relations which correspond to them whereas for expanded concepts, there are no knowledge relations describing them. By considering  the expanded concepts as input, we will be increasing the sparsity of the model and losing vital information that could be captured and this is essentially adding some amount of noise into the model. In order to avoid such scenario, we build masks to differentiate the given and expanded concept-sets within the injection mechanism. Attention is now computed only on the initial or given concept-set as is necessary.
    \end{itemize}

%%%%%%%%%%%%%%%%%%%%%%%%%%%%%%%%%%%%%%%%%%%%%%%
%%%%%%%%%%%%%%%%%%%%%%%%%%%%%%%%%%%%%%%%%%%%%%%

\begin{table*}[t]
  \centering
  \caption{Experimental results for knowledge injection through self-attention on knowledge relations}
%  \footnotesize
  \resizebox{\textwidth}{!}{
  \begin{tabular}{lllllllllll}
    \toprule
    \multirow{2}{*}{Experiment} &
    \multicolumn{4}{c}{BLEU ($\uparrow$)} &   \multicolumn{2}{c}{ROUGE ($\uparrow$)}                \\
    \cmidrule(lr){2-5}
    \cmidrule(lr){6-7}
     & BLEU-1 & BLEU-2 & BLEU-3 & BLEU-4 & ROUGE-L & ROUGE-2 & METEOR ($\uparrow$) & CIDEr ($\uparrow$) & SPICE ($\uparrow$) \\
    \midrule
   UniLM Baseline                                                                  & -                                                          & -                                                          & 38.3                                                         & 27.7                                                       & 43.87                                                           & 21.48                                                       & 29.7                                                  & 14.85                                                    & 30.2                                                                                                          \\

Attention + Knowledge selection                                           & 71.6                                                       & 52.5                                                       & 37.8                                                       & 27.0                                                       & 43.3                                                        & 21.67                                                       & 29.2                                                       & 14.57                                                        & 29.6                                    \\

Attention + Multi-hop                                         & 71.4                                                       & 53.3                                                       & 38.8                                                       & 28.1                                                       & \textbf{49.8}                                                        & \textbf{25.1}                                                        & 29.3                                                       & 14.78                                                      & 29.5                                    \\ 
Attention + Multi-hop + Best N Beam Scoring & 71.7 & 53.3  & 38.7  &    27.9 & 44.2   &     23.1 & 29.8 &    15.11     &  30.1  \\

Attention + Random subset selection & \textbf{72.4} & 53.2  & 38.1  &    27.2 & 43.72   &     22.45 & 29.8 &    14.92     &  30.2  \\

%\quad \quad \quad \quad \quad + Best N Beam Scoring & 71.6 & 53.2  & 38.1  &    27.2 & 43.72   &     22.45 & 29.8 &    14.92     &  30.2  \\
Attention + Prior subset selection &  71.8 & \textbf{53.5}  & \textbf{39.0}  &    \textbf{28.4} & 43.8   &     22.8 & 29.6 &    15.06     &  30.0
\\
Attention + Prior subset selection + Best N Beam Scoring & 71.9 & 53.4  & 38.7  &    27.9 & 44.4   &     23.43 & \textbf{30.1} &    \textbf{15.23}     &  \textbf{30.6}  \\

    \bottomrule
  \end{tabular}
  }
  \label{tab:experiments_better}
\end{table*}

\begin{table*}[t]
  \centering
  \caption{Experimental results for knowledge injection on Query Expansion}
%  \footnotesize
  \resizebox{\textwidth}{!}{
  \begin{tabular}{lllllllllll}
    \toprule
    \multirow{2}{*}{Experiment} &
    \multicolumn{4}{c}{BLEU ($\uparrow$)} &   \multicolumn{2}{c}{ROUGE ($\uparrow$)}                \\
    \cmidrule(lr){2-5}
    \cmidrule(lr){6-7}
     & BLEU-1 & BLEU-2 & BLEU-3 & BLEU-4 & ROUGE-L & ROUGE-2 & METEOR ($\uparrow$) & CIDEr ($\uparrow$) & SPICE ($\uparrow$) \\
    \midrule
   UniLM Baseline                                                                       & -                                                          & -                                                          & 38.3                                                         & 27.7                                                       & 43.87                                                           & 21.48                                                       & 29.7                                                  & 14.85                                                    & 30.2                                                                                                          \\ 

Concatenation + Query Expansion                                                           & 68.6                                                       & 49.1                                                       & 34.6                                                       & 24.5                                                       & 40.5                                                          & 19.4                                                           & 27.6                                                       & 13.26                                                    & 27.7     \\

%\quad \quad \quad \quad \quad + Best N Beam Scoring & 71.6 & 53.2  & 38.1  &    27.2 & 43.72   &     22.45 & 29.8 &    14.92     &  30.2  \\

Attention + Query Expansion + Multi-hop &  68.2 & 50.2 & 36.2 & 26.2  &  41.3 & 20.74  &   27.7  &     13.65  &    28.2\\
 
Attention + Query Expansion + Knowledge Selection & 67.9 & 49.3  & 35.3  &    25.3 &   41.27 &  20.22      & 27.5 &    13.50     &  27.6 \\

Attention + Constrained Query Expansion + Multi-hop & 69.6 & 51  & 36.5  &    26.1 &   42.0 & 20.5      & 28 &    13.82     &  28.3  \\

    \bottomrule
  \end{tabular}
  }
  \label{tab:experiments_worse}
\end{table*}

\subsection{Imposing Lexical Constraints}

Each concept in the dataset sample is of the format \textit{concept\_POS}, an example being \textit{drill\_N\#field\_N\#run\_V\#team\_N} where \textit{drill} is a concept with expected POS Tag \textit{Noun} while \textit{run} is another concept with expected POS Tag \textit{Verb}. We thus formulate the lexical constraint rules as follows:
\begin{itemize}
    \item Each of the given concepts should appear at least once in the generated sentence
    %\item None of the given concepts should be repeated more than once in the generated sentence
    \item Each of the given concepts should have the same POS tag in the generated sentence as given in the dataset
\end{itemize}

We experiment with concept and POS tag aware knowledge extraction as described previously (\nameref{subsection:extraction} section). The knowledge relations selected are constrained to select relations for each concept in the concept set, while rejecting any relations where the corresponding POS tag of the concept in the relation from \conceptnet~does not match the given POS tag. We are thus able to cover 99.57\% of all the unique concepts in the dataset. This constraint is also maintained during random subset selection, where subset of the relations is constrained to select at least one relation for each concept while determining the subset of relations.

\subsection{Best-\textit{N} Beam Scoring} We propose modifying constrained decoding by scanning the generations from the top N beams for their adherence to the constraints formulated above, N being a hyperparameter. We choose N = 4. For each of the N extracted sentences, we calculate the coverage score as product of \% of given concepts present in the generation * \% of concepts with correct POS Tag in the generation. The sentences with the highest coverage score are selected as generation. In case of a tie, the sentence with a higher beam score is selected.

%%%%%%%%%%%%%%%%%%%%%%%%%%%%%%%%%%%%%%%%%%%%%%%
%%%%%%%%%%%%%%%%%%%%%%%%%%%%%%%%%%%%%%%%%%%%%%%

\section{Experiments}
\label{section:experiments}
In this section, we evaluate the efficacy of our knowledge extraction and injection models on the \commongen~dataset with UniLM as our baseline language model and \conceptnet~as the knowledge graph.

%attention confusion matrix : Relative importance/weight given to each concept relation - how many useful relations were attended on

%Evaluations are performed on four extraction models: original Multi-hop, Multi-hop after Knowledge selection (length-based+Frequency-based) and Multi-hop after Constrained Randomization. These three models are constrained that we keep at least one relation for each input concept. We also experiment on unconstrained selection model: Multi-hop after unconstrained Randomization, which does pure randomization over all knowledge relations.

%It can be observed that two constrained selection method (Knowledge selection/Constrained Randomization) reduce 2.5 relations on average, while unconstrained Randomization reduce nearly half of the relations.
%In terms of the converge score, our multi-hop extraction is able to cover 99.57 percentage of input concepts on average, which would not change after constrained selection methods. For unconstrained random selection, the converge drops to 90.27 percentage. This is due to the chance that all relations of one input concept happen to be dropped. From constrained to unconstrained, we lost around 10 percentage of converge.

We evaluate the performance of the model for the task of lexically-constrained text generation w.r.t. human generations by calculating BLEU \cite{papineni2002bleu}, ROUGE \cite{lin2004rouge}, METEOR \cite{denkowski2014meteor}, CIDEr \cite{vedantam2015cider} and SPICE \cite{spice2016} scores.

As a baseline, we perform text generation using UniLM without any knowledge injection. We then experiment with two different strategies of knowledge injection: attention-based and concatenation based. 

\textbf{Concatenation based}: This is a baseline injection methodology inspired by rationale concatenation method by \cite{lin2020commongen}. We obtain expansion concepts using query expansion methodology as described previously and the tokens corresponding to these expansion concepts are then concatenated with the input concepts. Thus, for a given concept set \textit{run team drill field} and expansion set \textit{baseball sport football}, we send as input a concatenated list of concepts, \textit{run team drill field baseball sport football}. This method is referred to as \textit{Concatenation + Query Expansion} in later sections of the paper.

\textbf{Attention injection on knowledge relations}: Evaluations   are   performed   on   two major knowledge relation extraction models: (1) Multi-hop \textit{(Attention + Multi-hop)} and, (2) length and frequency-based Knowledge selection \textit{(Attention + Knowledge selection)}. Separately, we also experiment with performing Knowledge selection by selecting a subset of relations from multi-hop relation relations. For this method, the subset selection can be done in a randomized fashion by selecting the relations randomly with a probability of 0.5 (\textit{Attention + Random subset selection}) or by also associating a prior probability with each relation (\textit{Attention + Prior subset selection}), as explained in \textit{}{Knowledge Extraction Methodology} Section. In both these cases,  the subset selection is forced to select at least one relation for each input concept, wherever possible. 

\textbf{Attention injection on query expansion}: We experiment with  attention injection while incorporating the expansion concepts obtained from query expansion. We first experiment with all expansion concepts for both multi-hop \textit{(Attention + Query Expansion + Multi-hop)} and knowledge selection \textit{(Attention + Query Expansion + Knowledge Selection)} strategies for obtaining the knowledge relations on which the attention weights are calculated. Then we also experiment with the constrained version of Query Expansion with multi-hop \textit{(Attention + Constrained Query Expansion + Multi-hop)}, where each input sample contains at most 2 expansion concepts(with the highest frequencies). The threshold 2 is selected given that our target sentences across train/dev/test contain on average 2.5 additional verb/noun concepts. 

Further, we also integrate Best-\textit{N} Beam scoring with the two best performing models and evaluate the results.

%Using attention mask helps to distinguish between expansion concepts and the original inputs and only the original concepts are attended over. In this way, the expansion concepts provide extra knowledge while encoding but do not interfere with the knowledge injection.

%The next experiment on vanilla UniLM is to add additional concepts as an input. 

%As proposed in Injection Methodology section, we use a simple injection mechanism to inject the knowledge in the form of concept relations to the Vanilla UniLM model.

%The next experiment is a direct extension of the injection based architecture. We observed that the number of concept relations being input to the injection module was high. As pointed out in section \ref{knowledge selection} about Knowledge Selection, we reduce these relations.

%\textbf{As previously noted, this method does not show visible improvement however, we make use of an attention mask to distinguish between expansion concepts and the original inputs. Only the input concepts are attended over.}

\subsubsection{Assessing Effect of Lexical Constraints}

To evaluate the generated sentences w.r.t. the given lexical constraints, we calculate and report the following metrics:
 \begin{itemize}
     \item \% of given concepts missing in the text generated  
%     \item \% of given concepts that were present more than once in the text generated  
     \item \% of samples with any mismatching POS tag in the given input concept-set and generated text
 \end{itemize}

\begin{table*}[t]
  \centering
\begin{scriptsize}
  \caption{Analysis of lexical constraints}
 % \resizebox{\textwidth}{!}{
  \begin{tabular}{lllll}
   \toprule
Model  & Missing Concepts ($\downarrow$) & Mismatching POS ($\downarrow$) \\
    \midrule
    UniLM Baseline               & 42.9 &  22.45 \\  
 Attention +  Knowledge Selection      &  42.1   & 22.37\\
Attention +  Multi-hop         &  44.87     & 21.1 \\
   Attention +    Multi-hop  + Best N Beam Scoring &  28.12    & 21.64\\
     Attention + Random subset selection        & 39.7   & 21.81 \\
     Attention +     Prior subset selection  & 42.32  &  21.66\\     
 Attention + Prior subset selection + Best N Beam Scoring & \textbf{27.04}    & 21.69 \\
           \midrule
      Concatenation +       Query Expansion          &   61.2 & 21.4\\
 Attention +  Query Expansion + Multi-hop  &        60.8       & \textbf{19.31}    \\
  Attention +   Query Expansion + Knowledge Selection  & 59.5    &  20.42     \\
   Attention +   Constrained Query Expansion + Multi-hop & 59.5      & 20.9      \\

    \bottomrule
  \end{tabular}
  %}
   \label{tab:error_analysis}
   \end{scriptsize}
\end{table*}

% \begin{table*}[]
% \centering
% \caption{Coverage and error statistics}
% \resizebox{\textwidth}{!}{
% \begin{tabular}{|l|l|l|l|l|l|}
% \hline
% \multicolumn{1}{|c|}{\multirow{2}{*}{Experiments}} & \multicolumn{2}{l|}{Coverage statistics} & \multicolumn{3}{c|}{Error statistics}                     \\ \cline{2-6} 
% \multicolumn{1}{|c|}{}                             & Concepts              & POS             & Missing concepts & Repetitive concepts & Mismatching POS \\
% \hline
% UniLM                                      & 89.71 & 86.06 & 42.9 & 13.9  & 22.4  \\  \hline
% Query Expansion                          & 78.3  & 81.6  & 61.2 & 25.30 & 21.4  \\  \hline
% Knowledge Injection + Multi-hop                          & 88.21 & 84.59 & 45   & 18.7  & 19.57 \\  \hline
% Knowledge Injection + K.Select                           & 89.26 & 85.35 & 42.1 & 22.8  & 22.37 \\  \hline
% Variable Knowledge Injection + Query Expansions             & 78.39 & 81.65 &  60.8    &   16.21    & 19.23 \\  \hline
% Variable Knowledge Injection + Query Expansion + Knowledge Selection & 78.78 & 82.21 &  59.5    &  19.06     & 20.34 \\  \hline
% Knowledge  Injection + Knowledge Selection + Random sampling             & 86.62 & 90.01 & 39.7 & 30.4  & 21.8 \\ \hline
% \end{tabular}}   

%     \label{tab:my_label}
% \end{table*}

\begin{table*}[t]
  \centering
  \caption{Generated Sentences}
  \begin{footnotesize}
  \renewcommand{\arraystretch}{1.2}
  \resizebox{\textwidth}{!}{
  \begin{tabular}{p{0.25\textwidth}p{0.25\textwidth}p{0.25\textwidth}p{0.25\textwidth}}
    \toprule
    \multirow{2}{*}{Experiment} &
    \multicolumn{3}{c}{Generated Sentences}              \\
    \cmidrule(lr){2-4}
     & bed\_N comb\_V hair\_N sit\_V 
     & cover\_V front\_N mountain\_N short\_N wear\_V 
     & board\_N boat\_N ride\_V water\_N   \\
    \midrule
UniLM Baseline                                                          
& A woman with a comb and hair comb sits on a bed    
& person wearing shorts and long covers to the front of the mountain
& People board a boat and board a ride in the water\\ 

Attention + Knowledge Selection                                     
& A man with long hair sitting on a bed combing his hair 
& man wearing short shorts to the front of a mountain covered in snow
& A man rides a boat in the water to a ride on a boat .\\

Attention + Multi-hop                             
& A man sitting on a bed combing his hair with a comb .  
& man wearing a long sleeve shirt and shorts to the front of the mountain
& A man rides a boat in the water .\\ 

Attention + Multi-hop  + Best N Beam scoring                           & \textbf{A man sits on a bed and combs his hair . }
& man wearing a long sleeve shirt and shorts to cover the front of the mountain 
& A man rides a boat in the water with a boy on board .\\

Attention + Random Subset selection                        & A man with combs sits on a bed and combs his hair .    
& A man wearing a long sleeved shirt and shorts covered in snow covered mountains .
& A man rides a boat down a river and rides it into the water .\\

Attention + Prior Subset selection                        
& A man sits on a bed combing his hair with a comb . 
& A man wearing short shorts covered in snow covered mountains 
& A man rides a boat in the water . \\

Attention + Prior Subset selection  + Best N Beam scoring  & A man sits on a bed combing his hair with a comb
& \textbf{person wearing a short coat covered in snow on the front of mountain}
& \textbf{A man rides a boat in the water with a boy on board}\\

\midrule
   
Concatenation + Query Expansion                                     
& A man with comb and hair comb sits on a bed        
& A man wearing a long sleeved shirt is wearing a snow covered jacket
& A man sits on a boat and is riding a wooden boat in the water\\ 

Attention + Query Expansion + Multi-hop 
& A woman sits on a bed and combs her hair with a comb . 
& A man is wearing a long sleeved shirt to cover the front of a mountain        
& A man rides a boat on the water       \\

Attention + Query Expansion + Knowledge Selection    
& A woman sits on a bed and combs her hair with a comb . 
& A man is wearing a long sleeved shirt with a mountain covered in snow
& A man boards a boat and rides it down the water \\ 

Attention + Constrained Query Expansion + Multi-hop   
& A woman sitting on a bed combing her hair with a comb. 
& A man is wearing a long coat to cover his face .
& A man rides a boat in the water  \\ 
    \bottomrule
  \end{tabular}}
  \label{tab:generated_sentences}
  \end{footnotesize}
\end{table*}
\renewcommand{\arraystretch}{1}

\section{Results}
\label{section:results}

%i\jf{Separate `experiments' from `results'. `Experiments' is where we discuss the experimental design, explaining our decision to perform the experiments and why we think they will substantiate the claims we made above. `Results' is where we make reference to the tables and discuss the results of the experiments that were defined.}

We discuss the results obtained for the different proposed knowledge extraction and injection models in Tables \ref{tab:experiments_better} and \ref{tab:experiments_worse}. We find that \textit{Attention + Prior subset selection} model achieves the highest BLEU score. However, constrained decoding on this model using the Best-\textit{N} beam scoring (\textbf{\textit{Attention + Prior subset selection + Best N Beam Scoring}}) not only improves metrics such as METEOR, CIDEr and SPICE, but also achieved the least \% of missing concepts across all models, and is thus the best performing model. On the other hand, concatenation based methods perform poorly with decrease in model performance across metrics, as seen in Table \ref{tab:experiments_worse}. Our results show that attention based methods outperform other injection methodologies by huge margins.

Analyzing the sentences generated by baseline UniLM, we found that although the sentences appear correct grammatically, they lackred the required common sense knowledge to be meaningful. These sentences are still able to get a high BLEU score with a human generated sentence since the score is independent of the order of the n-grams. Although the scores for the proposed methods do not show much improvements, the generated sentences have a visible improvement in the commonsensical aspect of the sentence. We discuss this further with examples in the \nameref{section:discussion} section.

Poor performance of concatenation and query expansion method proves that feeding more inputs to the model might actually result in noise addition as the model has no way to differentiate between given and expanded concepts. These results further justify the need of an attention based injection mechanism. 

To do so, we incorporate the expansion concepts in the inputs as described in section: \emph{Injection methodologies - Attention for expansion concepts}. The use of expansion concepts provide extra knowledge while encoding but do not interfere with the knowledge injection. We tabulate the results of the masked injection model with and without knowledge selection. A small improvement in scores is seen, but the sentences seem to be much more natural and sensible. We also experiment with constraining the number of expansion concepts and see minor improvements in the performance scores, thus underlining the fact that too many expansion concepts may result in noise injection in the model. Instead, using a subset of expansion concepts, which are selected preferring high frequencies, can achieve better results.

\subsection{Constrained Decoding}
As can be seen in Table \ref{tab:error_analysis}, all our models perform better than the baseline UniLM model in terms of generating sentences with the correct POS tag, with more than 2 \% points improvement at a maximum. This can be directly attributed to knowledge selection, where relation selection was done in accordance with the given POS tag, wherever possible. 

We also see that Best-\textit{N} Beam Scoring is very effective in reducing the percentage of missing concepts, with upto 15\% improvements observed, whenever the method was applied. This can be attributed to the fact that beam search is biased towards choosing shorter sentences as product of token probabilities for a longer sentence is bound to result in a lower beam score. Larger number of concepts are naturally expected to result in longer sentences and exploring more beams enables choosing sentences with more concepts included, without compromising significantly on beam score. 

In the absence of beam search, we see that the percentage of missing concepts worsens drastically for concatenation based methods but shows slight improvements for attention based methods. In case of concatenation-based injection methods, this can be attributed to noise addition by expansion concepts, where the generated sentences often picked expansion concepts instead of the given concepts, hinting at model's inability to differentiate between the original and expanded concepts.

We also observed that across all the models, the repetition of concepts increased. Although this might not necessarily result in any metric/generation get better/worse, it is not in alignment with the usual human spoken English, where over-repetition of words is not expected.
 
%As seen in the above examples, we find that injecting commonsense knowledge into language model using attention is an incredibly effective methodology. Further regularization methods such as random subset selection from the knowledge set helps boost the performance and perform better on constrained decoding tasks while also enhancing the commonsensical nature of the sentences.

%%%%%%%%%%%%%%%%%%%%%%%%%%%%%%%%%%%%%%%%%%%%%%%
%%%%%%%%%%%%%%%%%%%%%%%%%%%%%%%%%%%%%%%%%%%%%%%

\section{Discussion}
\label{section:discussion}

%\input{Tables/extraction_experiment}
%\subsection{Qualitative Analysis}
We study the sentences generated by the baseline UniLM model and our proposed models and give few examples in Table \ref{tab:generated_sentences}. For the concept set \textit{bed\_N comb\_V hair\_N sit\_V}, we can see that the Unilm generated sentence covers all the given concepts but uses the word \textit{comb} as a noun instead of a verb, along with concept repetition. All our attention injection models were able to successfully maintain the lexical constraint of including all the concepts with the proper POS tag. We also see that the issue of repetition was rectified in the best sentences generated. 

In case of concept set \textit{cover\_V front\_N mountain\_N short\_N, wear\_V}, the sentence generated by baseline UniLM does not miss any concept but also does not make any sense. With the use of attention mechanism and external knowledge, the generated sentences improve on the commonsense nature of the sentences. In case of the concept set \textit{board\_N boat\_N ride\_V water\_N}, the baseline sentences not only lacked commonsense but also missed the concept \textit{board\_V}. We observe that Best-\textit{N} beam scoring on an attention-based model enabled including all the given concepts without compromising on the quality of the sentence generated.

% misses the concept \textit{laugh} while the attention injection enables inclusion of the missing concept in the sentence (\textit{A man is tickled and laughing while lying on a bed)} while also maintaining the commonsensical nature of the sentence. The most significant improvement is commonsense reasoning is seen for the third example \textit{catch\_V dog\_N frisbee\_N throw\_V}, where the UniLM generated sentence failed in commonsense reasoning \textit{A dog is throwing a frisbee into the air to catch it.}. Knowledge selection and attention injection alleviate this issue by injecting relevant relations such as \textit{["dog", "Synonym", "pawl", "IsA", "catch"]} and \textit{["frisbee", "RelatedTo", "disk", "UsedFor", "throw"]} which generates almost correct but significantly improved sentences, \textit{A man throws a frisbee to catch a dog} and \textit{A man throws a frisbee into a dog and catches it}.

%%%%%%%%%%%%%%%%%%%%%%%%%%%%%%%%%%%%%%%%%%%%%%%
%%%%%%%%%%%%%%%%%%%%%%%%%%%%%%%%%%%%%%%%%%%%%%%

\section{Future Work}
In our work, we experiment with various knowledge extraction methods and injection techniques. We observe that knowledge injection enables language models to perform better at text generation tasks that are lexically constrained. The improvement is visible in only a few examples and this motivates us to work towards improving our approach. As a future improvement, we plan to experiment further with constrained decoding where we plan to explore alternate methods to modify the beam score and give weightage to lexical constraints during the decoding process. We also wish to explore different attention mechanisms for knowledge injection. A self attention head based injection on the extracted concepts seems like a natural next step. Adding to that, defining a pre-training objective and pre-training the injection layer could help reduce the noise and generate much more meaningful sentences. \conceptnet~relations and the input concepts from the dataset have a PoS tag associated with them: thus, we also plan to explore PoS based encoding and decoding, where the unimodal latent representational of text takes into account a POS Tag based embedding. Moving aside from \conceptnet, we wish to see how the extraction techniques differ on various other knowledge graphs and come up with a generalized extraction mechanism.

\small
\bibliography{ref}

\end{document}